\documentclass[letterpaper,10pt,conference]{ieeeconf}

\usepackage{amsmath,amssymb,amsfonts}
\usepackage{graphicx}
\usepackage{subcaption}
\usepackage{booktabs}
\usepackage{cite}
\usepackage{stfloats}
\usepackage{tabularx}
\usepackage[hidelinks]{hyperref}
\hypersetup{
  pdftitle={Compositional Humanoid Manipulation through Isomorphic Teleoperation and Independent Skill Routing},
  pdfauthor={Yiming Wang, Peixian Yang, Lifeng Zhuo, Lai Wei, Xinling Du, Kun Song, Weiming Wang, Daolin Ma}
}

\IEEEoverridecommandlockouts
\title{\LARGE \bf Compositional Humanoid Manipulation through Isomorphic Teleoperation and Independent Skill Routing}

\author{Yiming Wang$^{1}$, Peixian Yang$^{2}$, Lifeng Zhuo$^{1}$, Lai Wei$^{3}$,\\
Xinling Du$^{4}$, Kun Song$^{2}$, Weiming Wang$^{1}$, and Daolin Ma$^{1}$%
\thanks{$^{1}$Shanghai Jiao Tong University.}%
\thanks{$^{2}$The University of Hong Kong.}%
\thanks{$^{3}$Xense Robotics.}%
\thanks{$^{4}$Fudan University.}%
}

\begin{document}

\maketitle
\thispagestyle{empty}
\pagestyle{empty}

\begin{abstract}
Humanoid manipulation requires reusable arm skills, yet demonstrating every possible skill combination is impractical. We present Humanoid Semantic Mixture-of-Experts Diffusion Policy (HS-MoDP), which composes separately demonstrated arm skills through independent semantic routing over a shared expert pool. This design preserves arm-specific skill control while sharing action-generation parameters across arms. A semantic scheduler integrates the learned policy with an existing locomotion controller through alternating manipulation and repositioning stages, extending the effective workspace and enabling skill reuse across location changes. To acquire demonstrations of whole-body contact manipulation, we develop G1Exo, an isomorphic exoskeleton that provides direct joint-level control of arm configurations. Experiments on a Unitree G1 show stronger skill composition than the baselines. G1Exo enables more reliable contact teleoperation than end-effector VR-IK, and its demonstrations improve contact-skill learning across four policy architectures. For HS-MoDP, switching from VR-IK to G1Exo demonstrations raises full cabinet workflow success from 35\% to 65\%. All project hardware designs and source code will be released as open source after the review process.
\end{abstract}

\section{Introduction}
\label{sec:introduction}

Humanoid manipulation uses dual-arm action, body contact, and mobility to interact with objects across an extended workspace~\cite{fu2024humanplus,ben2025homie,barreiros2025wholebody}. Gathering objects with the forearms or holding them against the torso requires control over arm configurations. Reusing separately demonstrated single-arm skills in new dual-arm combinations requires preserving each skill as the other arm's task changes. We study how to capture the configurations needed for contact manipulation and compose separately demonstrated skills during execution.

Vision-language-action (VLA) models learn robot behaviors from large-scale visual, language, and action data~\cite{kim2025openvla,pi05,grootn17}. Yet demonstrating every combination becomes increasingly costly: $m$ left-arm and $n$ right-arm skills admit $mn$ candidate pairings in a single stage. As analyzed in SkillVLA~\cite{zhai2026skillvla}, policies may retain demonstration-specific dependencies between arms in both predicted actions and internal policy representations. An arm's learned behavior may therefore depend on what the other arm did during demonstration. Successful execution of individual skills does not establish that each skill will be preserved when paired with a different one. We address this composition problem through separate skill conditioning and routing for each arm.

Capturing the configurations needed for body-contact manipulation is a complementary challenge. For a redundant seven-degree-of-freedom arm, an end-effector target does not uniquely determine elbow placement or forearm orientation~\cite{elias2024redundancy}. These configurations affect how the arms contact objects and support them against the torso. We develop G1Exo, a wearable exoskeleton kinematically isomorphic to the Unitree G1 arms. Direct joint-level mapping lets operators specify arm configurations, while a physical torso boundary provides a spatial reference for arm--torso interactions.

For skill reuse, we introduce Humanoid Semantic Mixture-of-Experts Diffusion Policy (HS-MoDP). Building on single-arm semantic interfaces and semantic expert routing~\cite{zhai2026skillvla,deng2026smodp}, HS-MoDP conditions two action streams on distinct skill representations and routes them independently through shared experts. Each stream receives its own wrist-camera and proprioceptive inputs alongside shared head-camera observations. This design retains scene context and shares action-generation parameters while limiting direct dependence on the other arm's local inputs. A frozen VLM schedules skills and coordinates the learned policy with an existing locomotion controller through alternating manipulation and repositioning stages.

Experiments on the Unitree G1 evaluate skill composition, whole-body contact teleoperation, and staged mobile manipulation. With equal demonstration counts, G1Exo improves contact-skill learning across four policies and increases HS-MoDP's full cabinet workflow success. Our contributions are threefold:
\begin{itemize}
    \item \textbf{G1Exo, a low-cost wearable isomorphic exoskeleton.} Joint-level control and a physical torso boundary support whole-body contact teleoperation and the collection of demonstrations that improve contact-skill learning across policy architectures.
    \item \textbf{HS-MoDP, a diffusion policy with independent routing and shared experts.} Arm-specific skill conditioning and routing support the composition of separately demonstrated skills into unseen dual-arm combinations.
    \item \textbf{A semantically scheduled system for staged humanoid loco-manipulation.} Skill scheduling integrates the learned policy with an existing locomotion controller to support staged manipulation with object-retaining repositioning.
\end{itemize}

We will release all project hardware designs and source code, including G1Exo and HS-MoDP, as open source after the review process.

\section{Related Work}

\subsection{Teleoperation and Demonstration Interfaces for Humanoid Manipulation}
\label{sec:related-work-demonstration}

Teleoperation interfaces determine which aspects of human motion are conveyed to a robot. Physical leader--follower systems provide demonstrations for bimanual and mobile manipulation~\cite{zhao2023aloha,fu2024mobile}, while human motion tracking and handheld grippers offer alternative collection interfaces~\cite{fu2024humanplus,chi2024universal}. For whole-arm contact, demonstrations must also specify configurations relevant to object contact.

Exoskeleton interfaces provide joint-level motion inputs for teleoperation and demonstration collection across embodiments~\cite{fang2024airexo,fang2025airexo,yang2024ace,zhong2025nuexo}. For humanoids, HOMIE integrates isomorphic arm control with body and hand interfaces for loco-manipulation~\cite{ben2025homie}, while HumanoidExo uses wearable recordings to support whole-body policy learning~\cite{zhong2025humanoidexo}. G1Exo focuses on directly specifying arm configurations and their spatial relationship to the torso during whole-body contact manipulation. Its low-cost wearable design combines joint-level arm control with a physical torso boundary, allowing operators to execute contact behaviors and provide corresponding demonstrations.

\subsection{Dual-arm Coordination and Compositional Skill Reuse}
Dual-arm policies model inter-arm dependence through joint action prediction~\cite{liu2025rdt1b}, intermediate representation exchange~\cite{lee2025interact,jiang2025decoupled}, or explicit role assignment~\cite{liu2025voxactb,bai2026pabicoop}. Structured latent representations and coordinated versus arm-specific MoE pathways further distinguish shared behavior from individual arm adjustments~\cite{wang2026covla,choi2026duallevel}. These mechanisms govern how the arms interact during execution. 
Composing separately learned skills additionally requires each arm's intended behavior to remain independently selectable.

Unimanual policy reuse has been explored through skill management, paired policy architectures, and energy-based coordination~\cite{lu2025anybimanual,im2026twinvla,song2026energyaction}. SkillVLA explicitly addresses skill entanglement and unseen left--right skill combinations~\cite{zhai2026skillvla}. It uses separate semantic instructions and low-level reasoning and action-generation streams for the two arms, with gated cross-arm communication to support cooperative behaviors. This establishes a formulation in which known skills can be invoked separately and recombined during execution. We study this formulation in humanoid joint-space manipulation, including concurrent skill execution in a changing shared workspace and manipulation involving arm--torso contact.

\subsection{Semantic Skill Conditioning with Mixture-of-Experts}
Sparse MoE policies use selective expert activation to share model capacity across tasks~\cite{shazeer2017moe,wang2025sdp}. The routing conditions determine how this capacity is organized: MoDE selects denoising experts according to the diffusion noise level~\cite{mode}, whereas MoE-ACT conditions routing on task context, visual observations, and proprioceptive states~\cite{guo2026moeact}. Skill-based approaches introduce additional structure, including primitive skill conditioning in diffusion policies and an orthogonal skill basis with sticky routing in SMP~\cite{gu2026skilldiffusion,hao2026abstracting}.

SMoDP connects expert selection to skill semantics through semantic routing, action-chunk consistency, and dual semantic alignment of skill representations and routing behavior~\cite{deng2026smodp}. It evaluates few-shot compositional adaptation with frozen experts on simulation benchmarks and bimanual multitask manipulation on ALOHA. We build on these routing and alignment mechanisms to support independently specified skills for concurrent arm execution. 
HS-MoDP gives each arm its own semantic condition and router within a shared diffusion transformer. 
In contrast to SkillVLA's separate single-arm action modules, the two streams access a common expert pool. Arm identity conditions the shared experts, allowing the model to distinguish the G1's mirrored arms while sharing action-generation parameters.

\section{G1Exo: Hardware for Humanoid-Aligned Whole-Body Teleoperation}
\label{sec:g1exo_hardware}

We developed G1Exo, a wearable bimanual teleoperation exoskeleton, to collect whole-body manipulation data for the Unitree G1 robot. Whole-body manipulation establishes contact with objects through body parts such as the arms and chest, using arm configurations and their positions relative to the torso to perform tasks such as holding and hugging~\cite{barreiros2025wholebody}. G1Exo uses a bimanual mechanism that is kinematically isomorphic to the G1 upper limbs. It maps exoskeleton joint angles directly to the corresponding robot joints, allowing the operator to specify arm configurations, including elbow placement and forearm orientation. Handle triggers control gripper opening and closing.

\begin{figure*}[!t]
    \centering
    \includegraphics[width=\textwidth]{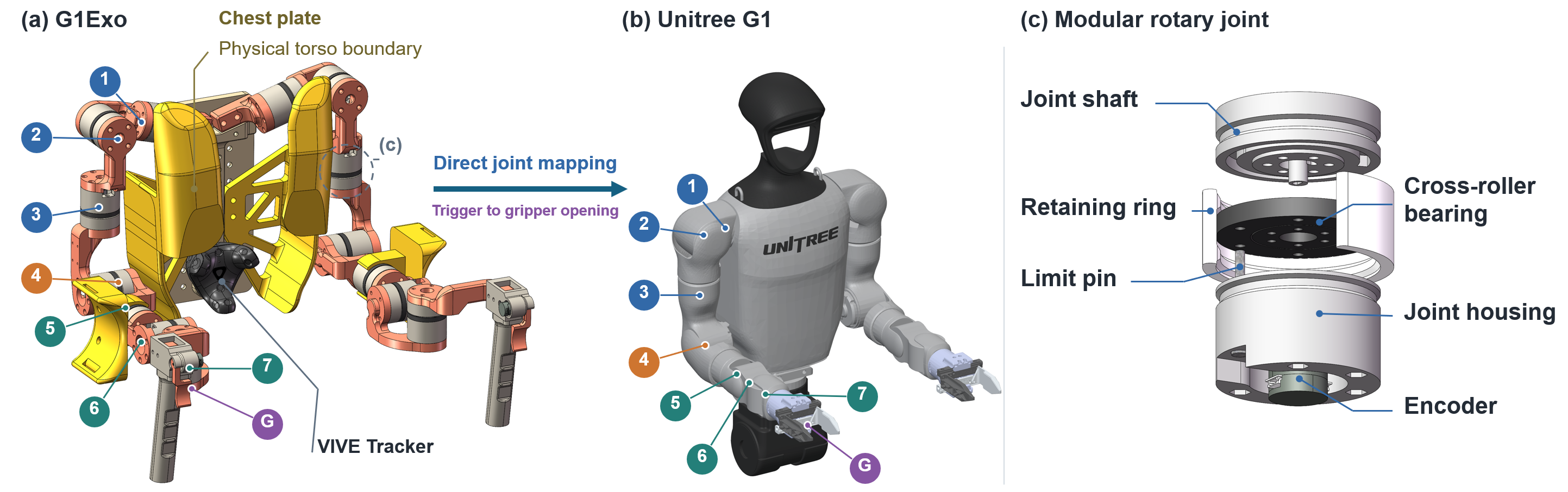}
    \caption{G1Exo hardware and joint mapping: (a) wearable exoskeleton; (b) corresponding joints on the Unitree G1; (c) exploded view of a modular rotary joint. Labels 1--7 denote shoulder pitch/roll/yaw, elbow, and wrist roll/pitch/yaw, respectively. G denotes gripper control via the handle trigger.}
    \label{fig:g1exo_hardware}
\end{figure*}

As shown in Fig.~\ref{fig:g1exo_hardware}(a--b), G1Exo consists of a base with a chest plate, two exoskeleton arms, and trigger handles. The arms match the kinematic parameters of the G1 upper limbs, including the arrangement and rotational limits of all 14 joints and the mounting positions of the arms relative to the torso. Each joint incorporates an angular encoder. After zero-position alignment, the measured angle serves as the motion input for the corresponding robot joint. Arm supports and straps secure the operator's arms to the exoskeleton, enabling direct control of the entire arm configuration through the operator's movements.

\begin{figure}[!t]
    \centering
    \includegraphics[width=0.86\columnwidth]{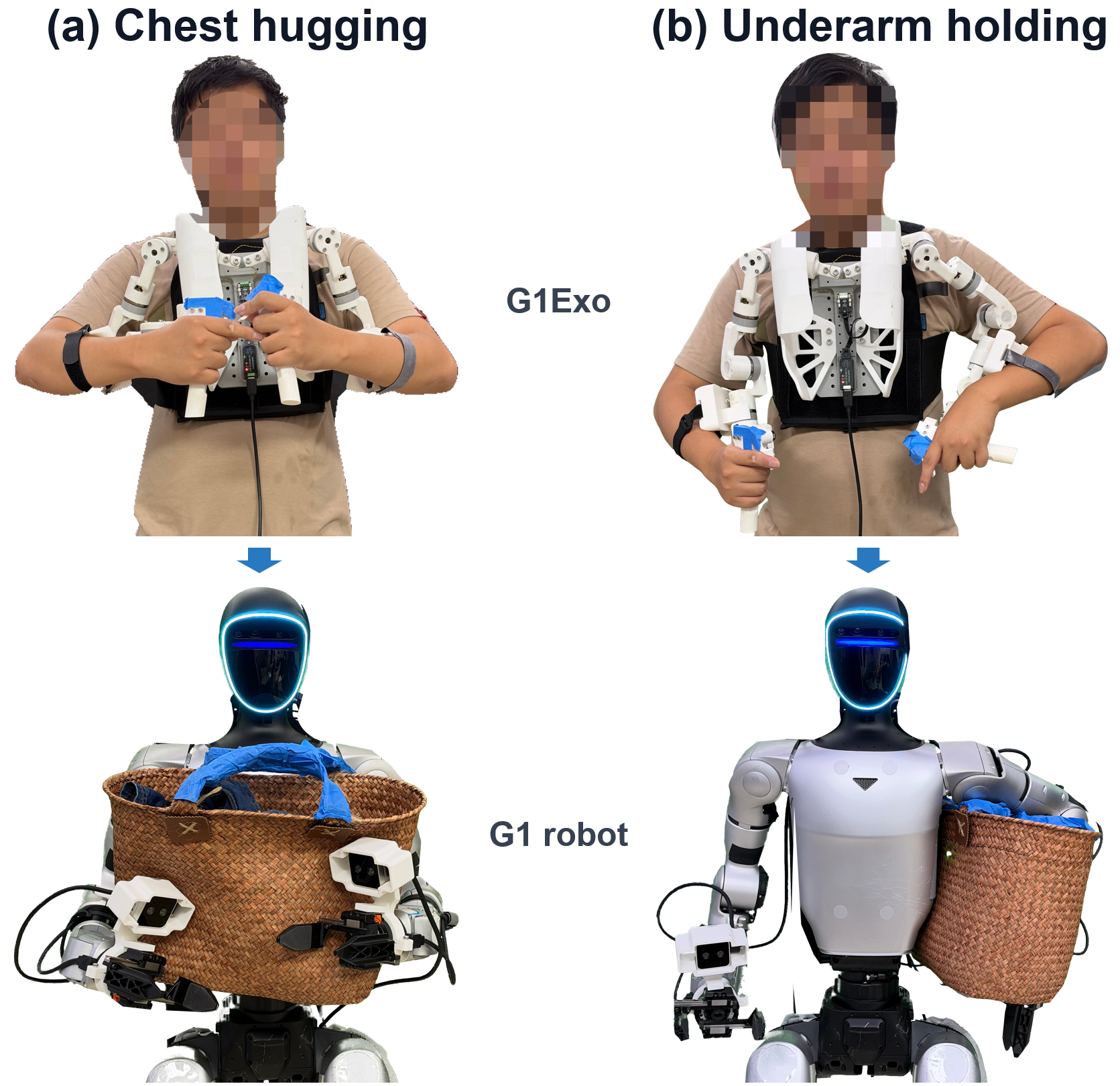}
    \caption{Photographs of whole-body contact manipulation teleoperated with G1Exo. The upper row shows the operator wearing G1Exo, and the lower row shows the corresponding G1 robot configurations: (a) hugging a basket against the chest and (b) holding a basket under the arm.}
    \label{fig:g1exo_actions}
\end{figure}

The chest plate provides a physical boundary corresponding to the robot's torso. As the arms move inward, the plate restricts further motion into the torso region, allowing the operator to perceive the spatial constraints between the arms and torso. A vest and hook-and-loop straps secure G1Exo closely to the operator's torso and reduce relative slipping during operation. The isomorphic arms and chest plate support chest hugging and underarm holding (Fig.~\ref{fig:g1exo_actions}), as well as bringing both arms inward around an object.

G1Exo uses reusable modular rotary joints, as shown in Fig.~\ref{fig:g1exo_hardware}(c). A cross-roller bearing supports relative rotation between the joint housing and shaft, a miniature rotary encoder measures the joint angle with 12-bit resolution, and a limit pin constrains rotation to the limits of the corresponding G1 joint. Links connect the joint modules to one another and to the base, forming the complete bimanual kinematic chains. Each terminal handle incorporates a trigger and an encoder of the same type, mapping trigger rotation to the robot gripper's opening width. The system contains 16 encoders: 14 measure arm joint angles, and two measure trigger inputs. All encoders communicate with the acquisition board via an RS-485 bus, with angle signals sampled at 100~Hz. A VIVE Tracker mounted on the base provides input for robot locomotion during mobile manipulation.

The structural components of G1Exo are 3D-printed from glass-fiber-reinforced ABS to balance wearable weight and structural strength. Mechanical hardware and electronic components are commercially available, mass-produced parts, keeping the total bill of materials (BOM) below USD~600. The design requires no specialized machining and can be reproduced in a laboratory using 3D printing and assembly of standard components.

\section{HS-MoDP: Single-Arm Routing for Compositional Dual-Arm Coordination}
\label{sec:hsmodp_method}

HS-MoDP (Humanoid Semantic Mixture-of-Experts Diffusion Policy) realizes single-to-dual-arm skill composition by treating separately learned arm behaviors as reusable skill primitives. Each arm independently selects experts from a shared pool according to its skill representation and combines their outputs using normalized routing weights, enabling previously unseen left–right combinations without requiring paired demonstrations.

\begin{figure*}[!t]
    \centering
    \includegraphics[width=\textwidth]{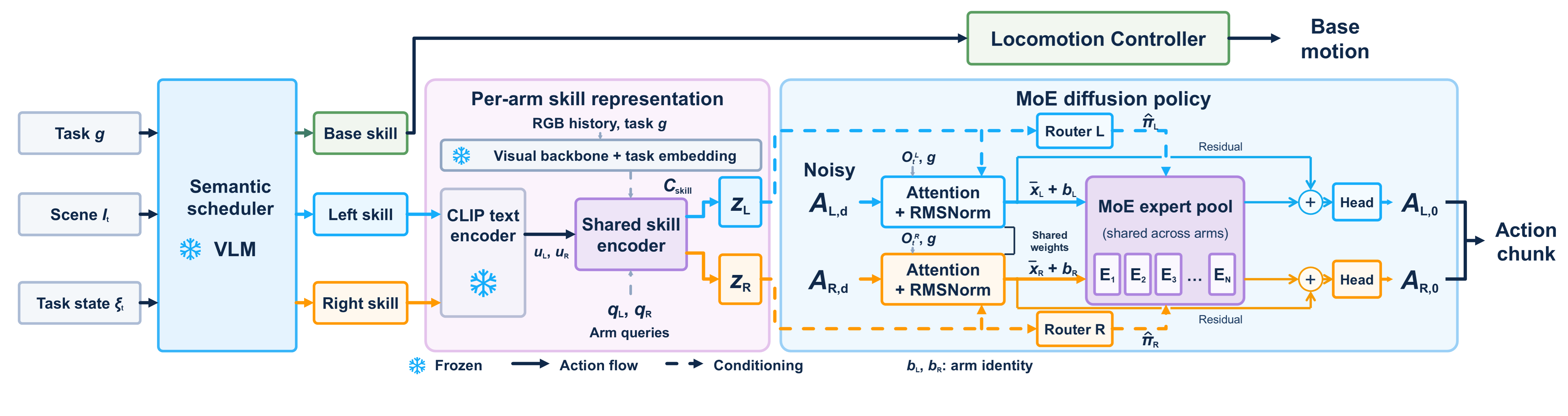}
    \caption{Overview of HS-MoDP. A frozen VLM-based semantic scheduler selects base and arm skills
    for locomotion and manipulation. Arm-specific skill representations condition the MoE diffusion policy,
    which generates separate action streams for the two arms. Solid and dashed arrows denote action flow
    and conditioning, respectively.}
    \label{fig:hsmodp_overview}
\end{figure*}

Fig.~\ref{fig:hsmodp_overview} presents the policy and its integration with a semantic
scheduler, which selects skills and connects the learned manipulation policy to an existing
locomotion controller.

\subsection{Problem Formulation}
\label{sec:hsmodp_coordination}

We consider humanoid manipulation tasks specified by a natural-language instruction $g$. For
skill composition, the objective is to execute concurrent left-right combinations of separately
demonstrated arm skills. At each replanning step $t$, the robot receives an observation history

\begin{equation}
    \mathcal O_t = \{\mathbf I_{t-T_o+1:t},\mathbf p_{t-T_o+1:t}\},
    \label{eq:hsmodp_observation}
\end{equation}
where $T_o$ is the observation horizon, $\mathbf I$ contains head and wrist RGB views,
and $\mathbf p$ contains measured arm-joint and gripper states.

Let $G$ denote the VLM-based semantic scheduler. Given the current images $\mathbf I_t$,
instruction $g$, and task-progress context $\xi_t$, it selects a skill configuration
\begin{equation}
    (b_t,s_{L,t},s_{R,t})
    = G(\mathbf I_t,g,\xi_t)
    \in \mathcal S_{\mathrm{base}}\times\mathcal S_L\times\mathcal S_R.
    \label{eq:hsmodp_scheduler}
\end{equation}
The sets $\mathcal S_{\mathrm{base}}$, $\mathcal S_L$, and $\mathcal S_R$ are the predefined
base, left-arm, and right-arm skill vocabularies, respectively. Here $b_t$ denotes a base
skill, including a stationary mode, and $s_{a,t}$ denotes the manipulation skill assigned
to arm $a\in\{L,R\}$. The text description of each selected arm skill provides the semantic
input to the conditioning module in Section~\ref{sec:hsmodp_skill_representation}.

During manipulation, the base remains stationary and HS-MoDP predicts an action chunk

\begin{equation}
    \mathbf A_t\in\mathbb R^{H\times16},\qquad
    \mathbf a_h=[\mathbf q_{L,h},\mathbf q_{R,h},g_{L,h},g_{R,h}],
    \label{eq:hsmodp_action}
\end{equation}

where $H$ is the prediction horizon, $\mathbf q_{a,h}\in\mathbb R^7$ is an arm-joint command,
and $g_{a,h}$ is a gripper command. The existing locomotion controller executes base skills.
Skill scheduling and transitions between manipulation and locomotion are described in
Section~\ref{sec:hsmodp_implementation}.

\subsection{Single-Arm Skill-Semantic Conditioning}
\label{sec:hsmodp_skill_representation}

Each arm receives a separate skill description. These descriptions specify the intended behavior
of each action stream and come from demonstration annotations during training. The two streams
share the task instruction and head-camera observations, while each stream uses its own
wrist-camera view and joint and gripper states. 
This asymmetric allocation preserves shared scene context while suppressing demonstration-specific cross-arm correlations in local sensing, thereby improving transfer from  single-arm demonstrations to dual-arm execution.

A frozen CLIP text encoder maps each skill description to an embedding $\mathbf u_a$, and a
frozen visual backbone encodes the camera observations. For arm $a$, the head-camera features,
its own wrist-camera features, and the task-instruction embedding form the context $\mathbf
C_{\mathrm{skill},a}$. A shared skill encoder $S_\phi$ combines an arm-specific query with the
projected skill embedding to extract relevant context:

\begin{equation}
    \mathbf z_a=S_\phi(\mathbf q_a+W_u\mathbf u_a,
                      \mathbf C_{\mathrm{skill},a}),\qquad a\in\{L,R\}.
    \label{eq:hsmodp_skill_encoder}
\end{equation}

The encoder applies cross-attention, layer normalization, and a feed-forward projection to
produce the $d_s$-dimensional representation $\mathbf z_a$. Its parameters are shared across
arms, while the learnable queries $\mathbf q_L$ and $\mathbf q_R$ distinguish the two arms.
During training, we independently set each arm's skill embedding to zero with probability $p_u$,
encouraging the encoder to also use visual and task context.

\subsection{Independent Routing with Shared Experts}
\label{sec:hsmodp_diffusion_transformer}

The diffusion transformer maintains a separate action stream and output head for each arm.
The two streams share attention and RMSNorm parameters while processing their action tokens
separately. Following the input allocation in Section~\ref{sec:hsmodp_skill_representation},
each stream is conditioned on its arm-specific observation history $\mathcal O_t^a$,
the task instruction $g$, and the skill representation $\mathbf z_a$.
The transformer denoises the dual-arm action chunk using these conditions, with $\mathbf z_a$
conditioning both action generation and expert selection.

We extend SMoDP's chunk-consistent semantic routing~\cite{deng2026smodp} with arm-specific
routers and a shared expert pool. Independent routing allows each skill to determine its expert
allocation, while shared experts reuse action-generation parameters across arms. At layer $l$,
an arm-specific projection $W_a^{(l)}$ and router $R_a^{(l)}$ compute expert probabilities:

\begin{equation}
    \boldsymbol\pi_a^{(l)}
    =\operatorname{softmax}\!\left(
      R_a^{(l)}(W_a^{(l)}\mathbf z_a)/\tau_r\right).
    \label{eq:hsmodp_router}
\end{equation}
where $\tau_r$ is the routing temperature. We sample $k$ experts during
training and select the top-$k$ experts at inference. The selected probabilities are
renormalized to sum to one, yielding sparse weights $\widehat\pi_{a,e}^{(l)}$. Both arms select
from the same $N$ feed-forward experts, with action-token update

\begin{equation}
    \mathbf x_{a,h}^{(l+1)}=
    \overline{\mathbf x}_{a,h}^{(l)}+
    \sum_{e=1}^{N}\widehat\pi_{a,e}^{(l)}
        E_e^{(l)}(\overline{\mathbf x}_{a,h}^{(l)}+\mathbf b_a^{(l)}).
    \label{eq:hsmodp_moe}
\end{equation}

The feature $\overline{\mathbf x}_{a,h}^{(l)}$ is RMS-normalized after the attention residual
update. The embedding $\mathbf b_a^{(l)}$ identifies the arm to the shared experts, allowing
them to distinguish the joint-space actions of the G1's mirrored arms. Expert parameters are
shared across arms within each layer and are independent across layers.
In Fig.~\ref{fig:hsmodp_overview}, the MoE block includes expert computation and per-arm weighted
aggregation, while the $+$ nodes denote residual addition.

Within each layer, all tokens in an arm's action chunk use the same routing weights. The
selected experts also remain fixed throughout a denoising rollout. Routing is recomputed at the next replanning step using
updated observations and skill conditions.

\subsection{Training Objective}
\label{sec:hsmodp_training}

We combine action denoising with semantic supervision that connects skill meaning to learned
representations and routing behavior. Adapting the diffusion and dual semantic alignment
objectives of SMoDP~\cite{deng2026smodp} to single-arm representations and routing gives

\begin{equation}
    \mathcal L=\mathcal L_{\mathrm{diff}}
      +\lambda_{\mathrm{inter}}\mathcal L_{\mathrm{inter}}
      +\lambda_{\mathrm{intra}}\mathcal L_{\mathrm{intra}}
      +\gamma\mathcal L_{\mathrm{bal}}+\delta\mathcal L_z.
    \label{eq:hsmodp_total_loss}
\end{equation}

The denoising loss $\mathcal L_{\mathrm{diff}}$ is the mean squared error for preconditioned
action denoising, averaged over supervised action elements. The inter-modal loss $\mathcal
L_{\mathrm{inter}}$ aligns learned skill representations with frozen skill-text embeddings. The
intra-modal loss $\mathcal L_{\mathrm{intra}}$ encourages similar routing for semantically
related skills. Both alignment losses use contrastive weights derived from skill-text
similarity.

For routing alignment, we use normalized, chunk-averaged sparse routing weights at each layer,
whereas SMoDP formulates this term using flattened routing logits. The load-balancing loss
$\mathcal L_{\mathrm{bal}}$ balances expert utilization, and the router z-loss $\mathcal L_z$
regularizes logit magnitudes. Alignment losses are computed separately for each arm and
averaged, using skill descriptions as the alignment labels. For demonstrations with only one
annotated arm, the other arm contributes neither action loss nor alignment supervision.

We train the skill encoder and diffusion transformer while keeping the visual and text backbones
frozen. The VLM-based semantic scheduler and the locomotion controller remain
fixed during training.

\subsection{Skill Scheduling and Closed-Loop Execution}
\label{sec:hsmodp_implementation}

At deployment, a frozen VLM selects skills from predefined vocabularies based on the task
instruction, scene observations, and task progress. Each skill
has a semantic label, an execution interface, and a completion condition, together with the
parameters required by its controller.

During manipulation, HS-MoDP uses the latest observation history and current arm-specific skills to
predict an action chunk. The robot executes a configured prefix before generating the next chunk
from new observations. This receding-horizon process updates action generation and expert
selection as the scene and selected skills change.

In the mobile cabinet task, manipulation-stage transitions are driven by VLM assessments of
task completion, aggregated over a recent observation window.
At the locomotion stage, the scheduler issues a predefined base repositioning command while the
arms hold their current pose. Manipulation resumes after the locomotion executor reports
completion, producing a sequence of manipulation and repositioning stages.

\section{Experiments}
\label{sec:experiments}

We investigate three questions: \textbf{Q1:} Can HS-MoDP compose separately demonstrated arm skills? \textbf{Q2:} Does G1Exo improve whole-body contact teleoperation and downstream policy learning over end-effector VR-IK? \textbf{Q3:} Can learned skills support manipulation with staged repositioning? Ablations examine independent routing, semantic alignment, and observation isolation.

\subsection{Experimental Setup}
\label{sec:exp_setup}

\textbf{Platform and configuration.} We use a Unitree G1 with two Dex1 grippers, a head-mounted RealSense D435i, and a RealSense D405 on each wrist. HS-MoDP predicts 16-dimensional actions (14 arm joints and two grippers) from two observation frames, using a 48-step horizon. Training and inference use a single NVIDIA RTX 5090 GPU.

\textbf{Baselines, datasets, and evaluation.} We compare HS-MoDP with the generalist policies $\pi_{0.5}$~\cite{pi05} and GR00T N1.7~\cite{grootn17}, and the MoE diffusion policy MoDE~\cite{mode}. For Table~\ref{tab:exp_autonomous}, each model is trained separately on G1Exo and VR-IK demonstrations. Both sources contain 100 demonstrations per whole-body contact task and 50 per cabinet skill, using the same collection and selection criteria. All other settings are matched across sources within each model. All methods receive identical task instructions and are evaluated in 20 trials per task and source under common success criteria, requiring all assigned subgoals or workflow stages within one trial.

\subsection{Unseen Dual-Arm Composition from Skill Primitives}
\label{sec:exp_skill_composition}

\subsubsection{Task Setup}

\begin{figure}[t]
\vspace*{5pt}
\centering
\includegraphics[width=\linewidth]{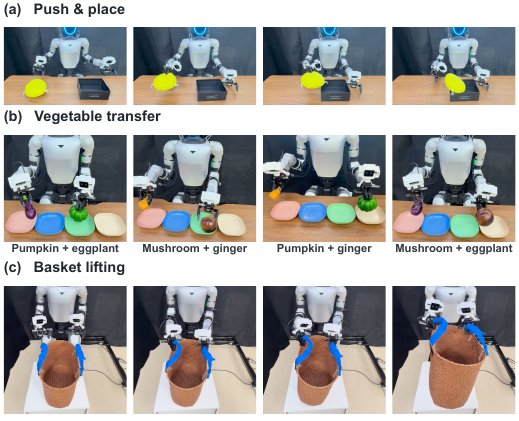}
\caption{Autonomous HS-MoDP execution: (a) box pushing with toy placement; (b) four left--right combinations; (c) bimanual basket lifting. Panels (a,c) show chronological sequences; (b) shows separate runs.}
\label{fig:exp_skill_composition}
\par\vspace*{-5pt}
\end{figure}

We test composition of separately demonstrated arm skills, using VLM-based planning for tabletop tidying and vegetable transfer (Fig.~\ref{fig:exp_skill_composition}). Tabletop tidying pairs left-arm box pushing toward the table center with right-arm placement of a toy or block into the box. Vegetable transfer uses four plates and tests four pairings of left-arm pumpkin/mushroom and right-arm eggplant/ginger transfers between designated source and target plates. 
For each single-arm skill primitive, we collect 50 demonstrations with one arm performing the operation and the other remaining stationary. The evaluated left--right combinations are absent from the demonstrations.

Basket lifting tests shared-load coordination: both hands must grasp the basket handles simultaneously and lift it stably. The left- and right-arm source skills are collected separately under the same protocol, with 50 demonstrations per skill and no demonstrations of their concurrent execution.

\subsubsection{Performance Results}

All methods could execute the source skills, but only HS-MoDP completed the six tabletop and vegetable combinations (Table~\ref{tab:exp_skill_composition}). 

HS-MoDP preserved the executability of the single-arm primitives while achieving a mean success rate of 73.3\% on unseen dual-arm combinations, indicating that its principal advantage lies in compositional transfer.

\begin{table}[t]
\vspace*{5pt}
\centering
\caption{Skill combination. Single-arm skill primitives and dual-arm combinations were evaluated separately.}
\label{tab:exp_skill_composition}
\footnotesize
\setlength{\tabcolsep}{3pt}
\renewcommand{\arraystretch}{1.08}
\begin{tabularx}{\linewidth}{@{}>{\raggedright\arraybackslash}Xrrrr@{}}
\toprule
Task & \multicolumn{4}{c}{Success (\%)} \\
\cmidrule(l){2-5}
& $\pi_{0.5}$ & GR00T & MoDE & Ours \\
\midrule
\multicolumn{5}{@{}l}{\textbf{Single-arm skill primitives}} \\
R: toy & 80 & 65 & 60 & 75 \\
R: block & 90 & 100 & 75 & 95 \\
L: box & 100 & 100 & 100 & 100 \\
L: pumpkin & 85 & 90 & 80 & 95 \\
L: mushroom & 85 & 80 & 70 & 85 \\
R: eggplant & 75 & 85 & 90 & 90 \\
R: ginger & 85 & 75 & 80 & 85 \\
L: basket lift & 100 & 100 & 100 & 100 \\
R: basket lift & 75 & 100 & 90 & 100 \\
\multicolumn{5}{@{}l}{\textbf{Dual-arm combinations}} \\
L: box + R: toy & 0 & 0 & 0 & 75 \\
L: box + R: block & 0 & 0 & 0 & 80 \\
L: pumpkin + R: eggplant & 0 & 0 & 0 & 85 \\
L: pumpkin + R: ginger & 0 & 0 & 0 & 65 \\
L: mushroom + R: eggplant & 0 & 0 & 0 & 60 \\
L: mushroom + R: ginger & 0 & 0 & 0 & 75 \\
L + R: basket lift & 5 & 0 & 0 & 65 \\
\bottomrule
\end{tabularx}
\par\smallskip\raggedright L/R: left/right arm; GR00T: GR00T N1.7.
\par\vspace*{-5pt}
\end{table}

For HS-MoDP, box pushing succeeded in every standalone and combined trial. Toy pick-and-place retained standalone performance during concurrent pushing, whereas block pick-and-place became less reliable. Successful vegetable combinations further demonstrated skill reuse across all left--right pairings.

Failed tabletop combinations showed interference with box pushing and arm oscillations without task progress. These observations suggest that concurrent execution can disrupt goal attainment even when the source skills are executable independently.

HS-MoDP outperformed the baselines in bimanual basket lifting, but remained less reliable than in either standalone arm skill. This gap identifies coordination under a shared load as a remaining challenge for skill composition.

\subsection{Contact-Aware Whole-Body Manipulation}
\subsubsection{Task Setup}

\begin{figure}[t]
\vspace*{5pt}
\centering
\includegraphics[width=\linewidth]{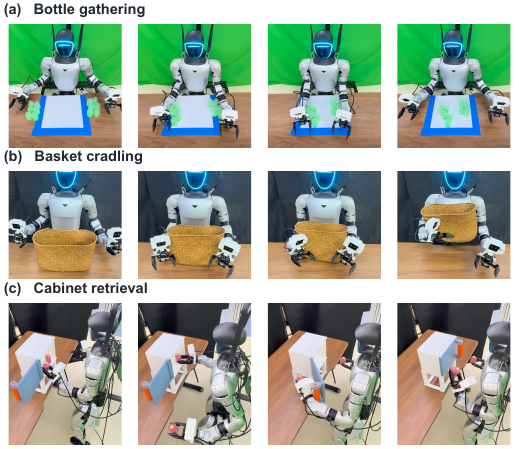}
\caption{Autonomous HS-MoDP execution: (a) forearm bottle gathering; (b) right-arm--torso basket cradling; (c) cabinet bottle retrieval with concurrent wrist-assisted door closing. Frames in each row proceed chronologically from left to right.}
\label{fig:exp_whole_body_mobile}
\par\vspace*{-5pt}
\end{figure}

Bottle gathering and handleless-basket cradling test whether G1Exo facilitates contact manipulation relative to VR-IK (Fig.~\ref{fig:exp_whole_body_mobile}(a,b)). Gathering sweeps 12 glass bottles, six per side, from random positions outside a central marked region into that region using forearm contact. Cradling uses both arms to pull the basket toward the body, then lifts it between the right arm and torso while maintaining contact.

\subsubsection{Teleoperation Interface Comparison}

We compare G1Exo's direct joint mapping with VR-IK, which converts two hand-controller poses into robot joint targets through inverse kinematics. Ten operators each performed ten trials per interface and task, giving 100 trials per condition. VR-IK used a Pico 4 headset, and operators practiced for ten minutes before evaluation.

\begin{figure}[t]
\vspace*{5pt}
\centering
\includegraphics[width=\linewidth]{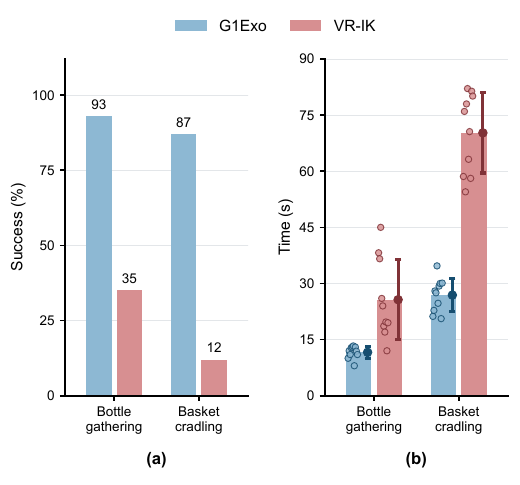}
\caption{Teleoperation on whole-body contact tasks. (a) Success rates. (b) Successful-trial completion times. Small points show participant means; bars and large points show their mean, and error bars indicate the sample standard deviation across participants.}
\label{fig:exp_teleoperation}
\par\vspace*{-5pt}
\end{figure}

Gathering succeeds when all bottles enter the target region upright through forearm contact within 60 s, without any bottle toppling during the trial. Cradling succeeds when the basket is lifted from the table and held stably between the right arm and torso for 10 s, with the basket remaining in contact with both throughout lifting and holding. Fig.~\ref{fig:exp_teleoperation} reports success rates over all trials and completion times averaged within each participant's successful trials, then equally across participants.

G1Exo achieved higher success rates and shorter mean completion times on both contact tasks. With VR-IK, the forearms often rose during gathering and toppled bottles; during cradling, unstable arm--torso support allowed the basket to slip. These failure patterns suggest that direct control of arm configuration helps maintain the contact required by both tasks.

\subsubsection{Policy Evaluation}

G1Exo demonstrations improved bottle gathering and basket cradling success over VR-IK across all four policies (Table~\ref{tab:exp_autonomous}). These gains extend G1Exo's benefits from contact teleoperation to autonomous policy learning.

\begin{table}[t]
\centering
\caption{Autonomous whole-body and mobile manipulation
success rates (\%) by demonstration source.}
\label{tab:exp_autonomous}
\footnotesize
\setlength{\tabcolsep}{1.5pt}
\renewcommand{\arraystretch}{1.0}
\begin{tabularx}{\linewidth}
{@{}>{\raggedright\arraybackslash}X*{4}{rr}@{}}
\toprule
Task
& \multicolumn{2}{c}{$\pi_{0.5}$}
& \multicolumn{2}{c}{GR00T}
& \multicolumn{2}{c}{MoDE}
& \multicolumn{2}{c}{Ours} \\
\cmidrule(lr){2-3}
\cmidrule(lr){4-5}
\cmidrule(lr){6-7}
\cmidrule(lr){8-9}
& G & \makebox[1.5em]{V}
& G & \makebox[1.5em]{V}
& G & \makebox[1.5em]{V}
& G & \makebox[1.5em]{V} \\
\midrule
\multicolumn{9}{@{}l}{\textbf{Individual skills}} \\
Bottle gathering
& 100 & 45 & 100 & 55 & 100 & 50& 100 &65 \\
Handleless-basket cradle
& 80 & 15& 70 & 10& 65 &10 & 85 & 15\\
L: retrieval
& 80 & 85& 65 &70 & 80 &75 & 80 & 75\\
R: retrieval
& 75 & 80 & 70 & 75 & 75 & 70 & 80 & 70\\
L: closing
& 90 & 45& 80 & 35& 75 &40 & 95 &45 \\
\multicolumn{9}{@{}l}{\textbf{Concurrent manipulation}} \\
R: retrieval + L: closing
& 0 &0 & 0 &0 & 0 &0 & 75 & 40\\
\multicolumn{9}{@{}l}{\textbf{Complete workflow}} \\
Full workflow
& 0 & 0& 0 &0 & 0 &0 & 65 &35 \\
\bottomrule
\end{tabularx}
\par\smallskip\raggedright
G/V: G1Exo/VR-IK training demonstrations;
20 trials per condition.
L/R: left/right arm; GR00T: GR00T N1.7;
Ours: HS-MoDP.
\end{table}

\subsection{Staged Humanoid Loco-Manipulation }
\subsubsection{Task Setup}

The cabinet task evaluates skill execution across manipulation and repositioning stages (Fig.~\ref{fig:exp_whole_body_mobile}(c)). The left hand retrieves a bottle from the cabinet and maintains its grasp during lateral repositioning. The right hand then retrieves another bottle while the left wrist closes the cabinet door, with both hands maintaining their grasps. VLM-based semantic scheduling coordinates the learned arm policy with an existing locomotion controller. Both data sources include separate demonstrations of left-hand retrieval, right-hand retrieval, and door closing. The concurrent skill combination is absent from both datasets.

\subsubsection{Performance Results}

Door closing is a contact-rich skill involving wrist--door interaction. G1Exo demonstrations improved its success across all policies (Table~\ref{tab:exp_autonomous}). Only HS-MoDP completed concurrent retrieval and closing and the full workflow with either source, and G1Exo data further improved its success.

\subsection{Ablation Study}
\label{sec:exp_ablation}

We examine independent routing, semantic alignment, and observation isolation using the box-pushing and toy-placement task and demonstration protocol in Section~\ref{sec:exp_skill_composition}. The box is both the left arm's manipulated object and the right arm's placement target, testing skill composition in a changing shared workspace. Each configuration is evaluated separately on both unimanual skill primitives and their combination, with 20 trials per setting and matched initial-scene distributions. Standalone success requires the box to reach the table center or the toy to be grasped and placed inside the box, respectively. Combined success requires both goals within one episode. The variants below otherwise follow the full model's configuration.

\textbf{Coupled routing.} We retain both arm-specific routers and average the dense routing probabilities of arms assigned active manipulation skills at each layer. One shared subset of $k$ experts is sampled during training or selected by top-$k$ at inference, with identical renormalized weights applied to both streams. Routing remains consistent across each action chunk and denoising rollout.

\textbf{Without semantic alignment.} We set $\lambda_{\mathrm{inter}}=\lambda_{\mathrm{intra}}=0$ while retaining skill-text conditioning, action denoising, load balancing, and the router z-loss.

\textbf{Unified observations.} Both skill encoding and diffusion action generation receive the head and both wrist-camera views, and each action stream receives the complete dual-arm joint and gripper states. Single-arm skill conditions, independent routing, and shared experts are retained.

\begin{figure}[t]
\vspace*{5pt}
\centering
\includegraphics[width=82mm]{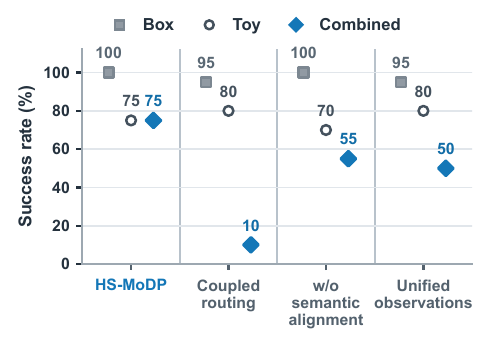}
\caption{Ablation on box pushing and toy pick-and-place. Box and Toy denote separately evaluated unimanual skill primitives; Combined denotes their concurrent execution.}
\label{fig:exp_ablation}
\par\vspace*{-5pt}
\end{figure}

All ablations retained comparable standalone performance but reduced combination success (Fig.~\ref{fig:exp_ablation}). Coupled routing caused the largest decline, with failures to release the toy or premature release during placement, while box pushing was largely preserved. Without semantic alignment, the toy arm often failed to adjust placement when the box stopped short. Unified observations produced repeated corrections and oscillations, consistent with a mismatch between separately collected demonstrations and the cross-arm inputs encountered during concurrent execution. These patterns support independent expert allocation, semantic alignment, and local observation isolation for skill composition.

\section{Conclusion}
\label{sec:conclusion}

We presented HS-MoDP for composing separately demonstrated arm skills through independent semantic routing over shared experts, and G1Exo, a low-cost wearable isomorphic exoskeleton for joint-level control and demonstration collection. A VLM-based scheduler integrates the learned policy with an existing locomotion controller across manipulation and repositioning stages. Experiments on the Unitree G1 showed stronger skill composition than the baselines, with ablations supporting independent routing, semantic alignment, and observation isolation. G1Exo improved contact teleoperation over VR-IK, and its demonstrations improved contact-skill learning across four policy architectures. HS-MoDP composed skills using either demonstration source, with G1Exo data further improving concurrent manipulation and full cabinet workflow success.

Future work will expand the skill repertoire and explore contact-aware coordination for simultaneous manipulation and locomotion.

{\footnotesize
\bibliographystyle{IEEEtran}
\bibliography{related_work_v3}
}

\end{document}